\documentclass[journal]{IEEEtran}

\usepackage{cite}
\usepackage{bm}
\usepackage[linesnumbered,ruled]{algorithm2e}
\usepackage{graphicx}
\usepackage{subfigure}
\usepackage{lineno,hyperref}
\usepackage{amsmath}
\usepackage{pifont}
\usepackage{epstopdf}
\usepackage{multirow}
\usepackage{amsthm,amsmath,amsfonts,amssymb,bm}
\usepackage{array}
\usepackage{booktabs}
\usepackage{makecell}
\usepackage[all]{xy}

\modulolinenumbers[5]

\ifCLASSINFOpdf
\else
\fi

\begin{document}

\title{Labeled Subgraph Entropy Kernel}

\author{Chengyu~Sun, Xing~Ai, Zhihong~Zhang$^{\ast}$, and~Edwin~R~Hancock,~\IEEEmembership{Fellow,~IEEE}
% <-this % stops a space
    % \IEEEcompsocitemizethanks{
    %     \IEEEcompsocthanksitem Chengyu Sun are with School of Informatics, Xiamen University, Xiamen, Fujian, China.
    %     \protect\\ E-mail: 24320191152507@stu.xmu.edu.cn, 30920201153942@stu.xmu.edu.cn
    %     \IEEEcompsocthanksitem Corresponding author: Zhihong Zhang is with School of Informatics, Xiamen University, Xiamen, Fujian, China.
    %     \protect\\ E-mail: zhihong@xmu.edu.cn
    %     \IEEEcompsocthanksitem Edwin R. Hancock is with University of York, York, UK.  
    %     \protect\\ E-mail: edwin.hancock@york.ac.uk}
% <-this % stops an unwanted space
}


\markboth{Journal of \LaTeX\ Class Files,~Vol.~14, No.~8, August~2015}%
{Shell \MakeLowercase{\textit{et al.}}: Bare Demo of IEEEtran.cls for IEEE Journals}



\maketitle

\begin{abstract}
In recent years, kernel methods are widespread in tasks of similarity measuring. Specifically, graph kernels are widely used in fields of bioinformatics, chemistry and financial data analysis. However, existing methods, especially entropy based graph kernels are subject to large computational complexity and the negligence of node-level information. In this paper, we propose a novel labeled subgraph entropy graph kernel, which performs well in structural similarity assessment. We design a dynamic programming subgraph enumeration algorithm, which effectively reduces the time complexity. Specially, we propose labeled subgraph, which enriches substructure topology with semantic information. Analogizing the cluster expansion process of gas cluster in statistical mechanics, we re-derive the partition function and calculate the global graph entropy to characterize the network. In order to test our method, we apply several real-world datasets and assess the effects in different tasks. To capture more experiment details, we quantitatively and qualitatively analyze the contribution of different topology structures. Experimental results successfully demonstrate the effectiveness of our method which outperforms several state-of-the-art methods.
\end{abstract}

\IEEEpeerreviewmaketitle
\section{Introduction}
For the past few years, research on learning non-Euclidean space data has become a central topic of pattern recognition. Graphs can describe numerous problems efficiently because of their competence in characterizing structured data in non-Euclidean space. In the field of graph data analysis, graph comparison has been an important branch which has been applied in many areas such as bioinformatics, chemistry, and sociology. Studies on graph comparison have been divided into the following three groups: Set-based approaches, Frequent subgraph mining (FSM), and Graph kernels. Set-based approaches represent a graph as a set of edges and nodes, then measure the similarity between pairs of sets. Despite feasible computational complexity, they neglect the inherent topology information. Frequent subgraph mining algorithms aim to detect subgraphs that occur frequently and select the discriminative subgraphs. Borgelt et.al\cite{borgelt2002mining} present an algorithm to find fragments in a set of molecules that help to discriminate between different classes of activity in a drug discovery context. Deshpande et.al\cite{deshpande2005frequent} classified chemical compounds by considering frequent topological and geometric substructures present in the data set. Unfortunately, their computational complexity scales exponentially with graph size.

Striving for balance between richer information and lower complexity, graph kernels represent an attractive middle ground. The idea of constructing kernels between graphs was first proposed by G{\"a}rtner\cite{gartner2003graph} and extended by Borgwardt \cite{borgwardt2005protein}. Graph kernels not only respect graph topology, but also restrict the computational complexity in polynomial time\cite{shervashidze2009efficient} by switching the problem from the vectorial representation of graph to a representation of similarity\cite{xu2021deep}. Graph kernels are composed of a learning algorithm and the kernel function, which is an asymmetric positive semi-definite function that measures the similarity between examples. With high efficiency in structural data comparing, graph kernels have been applied heavily in fields include bioinformatics\cite{airola2008graph}, chemistry\cite{ralaivola2005graph} and financial data analysis\cite{cui2016p2p}. 

In fact, many of the complex networks that occur in nature can be succinctly described with simple statistical subgraphs\cite{strogatz2001exploring}. For example, researches show that graphlets and motifs perform specific functional roles in a large network structure\cite{milo2002network}. Motifs are recurring patterns that can be used in representation of of complex structure. Indeed, motifs reflect not only the structural properties of a network, but can also capture its functional properties too. Similarly, graphlets are small connected non-isomorphic patterns recurring in real-world networks. Their frequencies are statistically significant in biology data analysis\cite{milenkovic2008uncovering}, protein function prediction \cite{shervashidze2009efficient}, network alignment\cite{milenkovic2010optimal}, and phylogeny\cite{kuchaiev2010topological}. While graphlets have witnessed tremendous success and impact in a variety of domains, there has yet to be an analysis for combining topology and semantic information. Moreover, most of the chemical applications are not only interested in finding frequent graphs but in identifying significant patterns which seldom occur. Thus, it is an interesting work to identify which substructure has the cardinal influence on the graph classification tasks\cite{ramraj2015frequent}.

For the sake of exploring the deep relationship between local property and global feature, tools from statistical mechanics provide a convenient way to characterize the network structure\cite{albert2002statistical}. This task requires an understanding of the basic structural elements constituting the graph and the processes which give rise to them from a microscope point of view\cite{alon2007network}. Zhang et al.\cite{zhang2020graph} uses the motif content and cluster expansion to compute the thermodynamic entropy and conducts numerical experiments which prove that network motifs can be regarded as basic elements with well-defined information processing functions. The cluster expansion is a powerful computational tool that can be used to express the partition function in terms of an approximating series\cite{salpeter1958mayer}. It allows us to write the grand-canonical thermodynamic potential as a convergent perturbation over the interactions between particles. Commencing from the general principles of perturbation theory for particle systems, the cluster expansion allows us to understand complex systems of interactions in terms of the motif in a diagrammatic expansion of the partition function. Existing graph cluster expansion applications are based on motif and ignore the node-level (semantic) information, so it is necessary to improve the process with more comprehensive statistical element.

In this paper, we propose a entropy-based graph kernel method, the main contributions of this work are threefold. 
\begin{enumerate}[(1)]
    \item We design a breadth-first subgraph mining algorithm with a dynamic programming process, which calculates the number of predetermined labeled structures. Our algorithm achieves lower time complexity when counting subgraphs with large scale.
    \item We re-derived the partition function with the broader term "subgraph" instead of graphlet, then we calculated the global entropy. By considering the labeled subgraph, this statistical variable could reflect the randomness of a graph using both semantic and topology information. 
    \item We propose a novel graph embedding process based on the existing framework of frequent subgraph mining methods. On this basis, we propose the Labeled Subgraph Entropy Kernel (LSEK) to measure the similarity between graphs, and verify the symmetry and positive definitiveness of LSEK.
\end{enumerate}

Our subgraph entropy kernel is performed in multiple types of application scenarios, such as biology and financial dataset. The qualitative experiment results show that our method has a powerful capacity in extracting implicit structural information in financial networks. The quantitative experiment shows that our kernel outperforms several state-of-the-art graph classification algorithms in real-world biochemistry datasets.

\section{related work}
\subsection{graph kernel}
Kernel methods are strong learning algorithms that can be effortlessly used for measuring the similarity between structured objects $x$ and $x’$ with a kernel function $k(x,x’):\rightarrow\langle \varphi\cdot\varphi'\rangle$, as $k$ corresponds to an inner product in Reproducing Kernel Hilbert Space (RKHS)\cite{scholkopf2002learning}. When $x$ and $x’$ are individual graphs, the challenge is to construct a kernel that captures the inherent relevance between graph structures. It is remarkable that R-convolution\cite{haussler1999convolution} is a general framework for handling discrete objects, which key idea is to recursively decompose objects into “atomic” sub-structures and define valid local kernels between them. In fact, most graph kernels have been defined with a similar idea, which focuses on deconstructing two distinct objects and comparing some simpler substructures. 

Graph kernels are generally classified into three broad categories: path/random walk based kernels\cite{gartner2003graph}\cite{borgwardt2005shortest}, subtree based kernels\cite{ramon2003expressivity}\cite{shervashidze2011weisfeiler} and subgraph based kernels\cite{horvath2004cyclic}\cite{yanardag2015deep}. Random Walk Kernels\cite{gartner2003graph} performs random walks on both object graph, and count the number of matching walks. Shortest Path Kernels\cite{borgwardt2005shortest} are not only applicable to a wide range of graphs but computable in polynomial time. When it comes to subtree kernel, the famous Weisfeiler-Lehman subtree kernel\cite{shervashidze2011weisfeiler} belongs to this family. The key idea is to iterate over each vertex and its neighbors in order to create a multiset label and then simply count the co-occurrences of labels in both graphs. Graphlets have been proved to be effective for the intuitive and meaningful characterization of networks at both the global macro-level as well as the local micro-level. Frequently occurring patterns within a network structure, such as motifs and graphlets, have been proved to be effective for intuitive and meaningful characterization of networks at both the global macro-level as well as the local micro-level\cite{ahmed2015icdm}. By using these features, Shervashidze et al.\cite{shervashidze2009efficient} define a kernel based on the distribution of graphlets for unlabeled graphs and proposed an efficient counting algorithm with two theoretically grounded speedup schemes. On this basis, Yanardag et al.\cite{yanardag2015deep} present Deep Graph Kernels, a unified framework to learn latent representations of sub-structures. However, few studies have focused on subgraph-based kernels with discrete node labels, so it is an interesting work to combine topology and semantic information.

\subsection{Sub structures and Cluster expansion}
Numerous studies have shown that network substructures, such as motifs\cite{milo2002network} \cite{shen2002network} and graphlets\cite{ahmed2015icdm}, can be used in the representation of more complex structure. For example, in the field of biochemistry, network motifs have been implicated in signalling\cite{awan2007regulatory} and neuronal activities\cite{varshney2011structural}. In social science, graphlets are widely adopted in sociometric studies\cite{holland1976local} \cite{frank1988triad}. To embark on the analysis of statistical features from the substructures perspective, thermodynamic characterizations such as entropy provide a convenient route to succinctly describe the network statistics\cite{wang2017spin}. 

From a statistical mechanics point of view, networks can be taken analogous to interacting particles systems. On this basis, the cluster expansion provides deep insights into network behavior related to the occurrence of different motifs. The cluster expansion is usually a power series expansion of the partition function and describes the pattern of interactions in a system with a large number of particles. Mayer et al.\cite{salpeter1958mayer} carried out a systematic study of real gases obeying classical statistics. Lee et al.\cite{lee1957many} have explored these ideas in a real-world application. In recent years, several studies have focused on entropy to represent the statistical topology of networks. Zhang et al.\cite{zhang2020graph} find that the motif entropy for financial stock market networks is sensitive to the variance in network structure. Chen et al.\cite{chen2021thermodynamic} present a novel thermodynamically based analysis method for directed networks. Existing research is confined to network motif-based entropy and ignores node-level granularity. Hence they are not accessible to general subgraph-based applications.

The remainder of this paper is organized as follows. In Section 3, we present the detailed process of our Labeled Subgraph Entropy Kernel. In Section 4, we introduce the experimental design and corresponding results. Finally, in section 5, we conclude our method and put forward directions for future work.

\section{Labeled Subgraph Entropy Kernel}
As mentioned above, our work is consists of three parts: subgraph counting algorithm, subgraph based entropy calculating and kernel definition. In this section, we provide a detailed description of our statistical process. We propose a network statistic element called Labeled Graphlet and design the corresponding searching algorithm. Next, we demonstrate the re-derivation process of subgraph-based cluster expansion and calculate the global entropy of the network. Finally, we define the subgaph entropy kernel and prove its corresponding required properties. In order to make our description clearer, we demonstrate the whole process of calculating subgraph entropy embedding in Algorithm \ref{alg:entropy}.

\iffalse
\begin{table}
\centering
\begin{tabular}{lll}
\hline
Symbol  & Definition  \\
\hline
$G$&Graph set\\
$GL$&Graph labels\\
$G_i$&The $i$th graph in $G$\\
$L$ & Node label set \\
$l$ & $|L|$,the number of node label types\\
$t$ & Number of designated graphlet topology types\\
$N$ & $|G|$, the size of graph-set $G$\\
$n$&Number of nodes in $G_i$ \\

\hline
\end{tabular}
\caption{Important notation used in this paper and their descriptions.}
\label{tab:plain}
\end{table}
\fi

\begin{algorithm}[htb]
\label{alg:entropy}
\caption{Subgraph Entropy Algorithm}
\KwIn{Adjacency matrix $adj$ and node label set $L$ of individual graph $G$, specific subgraph type $v$.}
\KwOut{Thermodynamics entropy value $S$ of $G$ which measured by $v$-th subgraph}
\textbf{Initialize super parameters}: inverse temperature $\beta$, scale parameter $\sigma$, the subgraph node number $l_v$ and edge number $d_v$ of $v-$th subgraph;\\
\textbf{Graphlet counting}: calculate the occurrence frequency $n=\{n_1,n_2...n_L\}$ of $v-$th subgraph starting from each node label within $G$;\\
\textbf{Edge configuration integral}: compute edge configuration integral $\zeta_v$ with formula(3);\\
\textbf{Partition function}: compute Subgraph configuration integral $q_v$ and the partition function $z_v$ of $v-$th labeled graphlet with formula(1) and formula(2);\\
\textbf{Subgraph entropy}: Calculate the subgrpah entropy $S_v$ of $v-$th  graphlet;\\
\textbf{Return $S_v$}
\end{algorithm}

\subsection{Labeled Subgraph Counting}
\begin{figure*}
\centering
\subfigure[Difference between motif and graphlet]{
\label{fig:graphlet_vs_motif}
\includegraphics[width=0.4\textwidth]{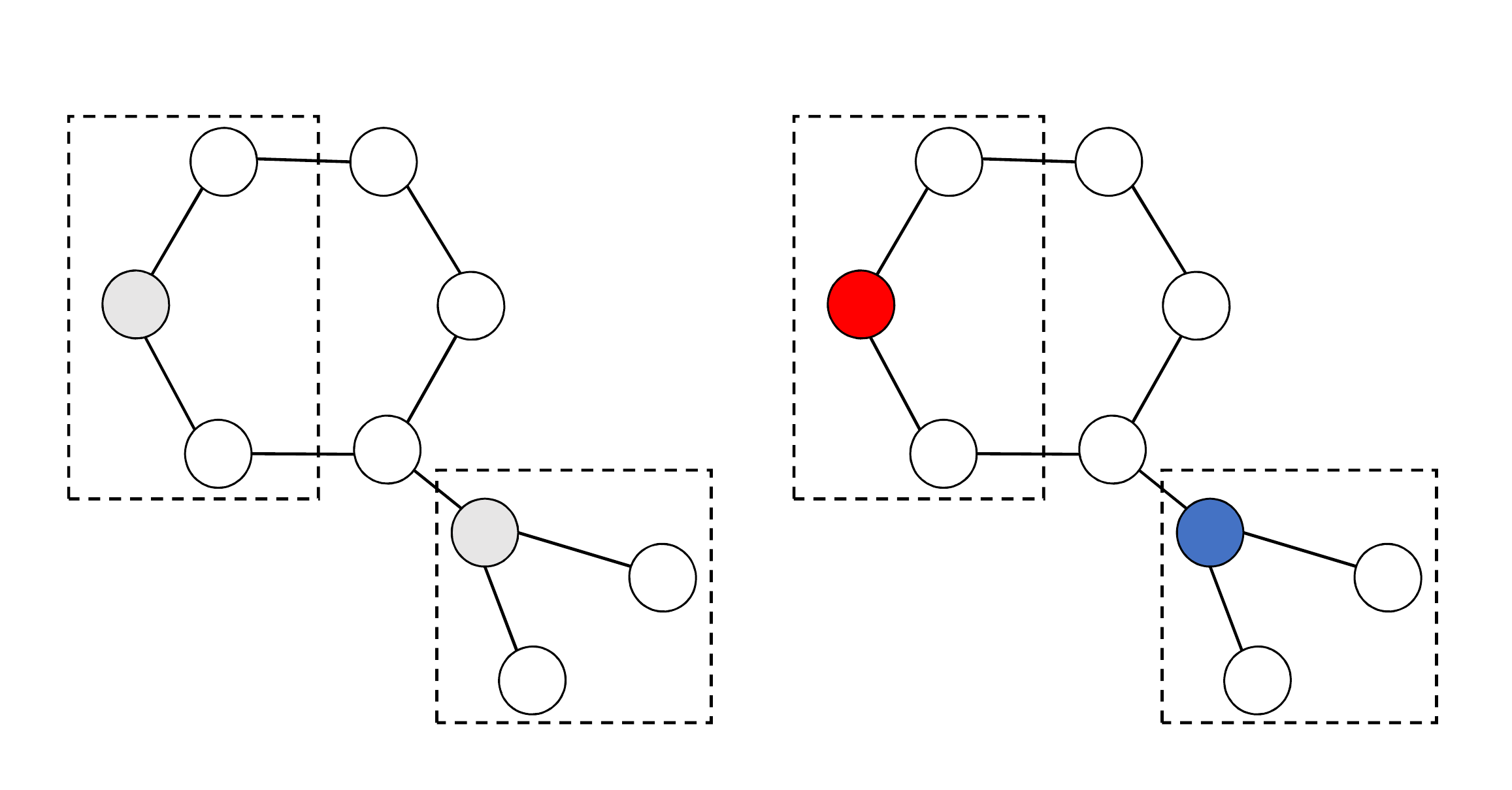}
}
\quad
\subfigure[All twelve types of the structures mentioned in this paper.]{
\label{fig:graphlet_types}
\includegraphics[width=0.4\textwidth]{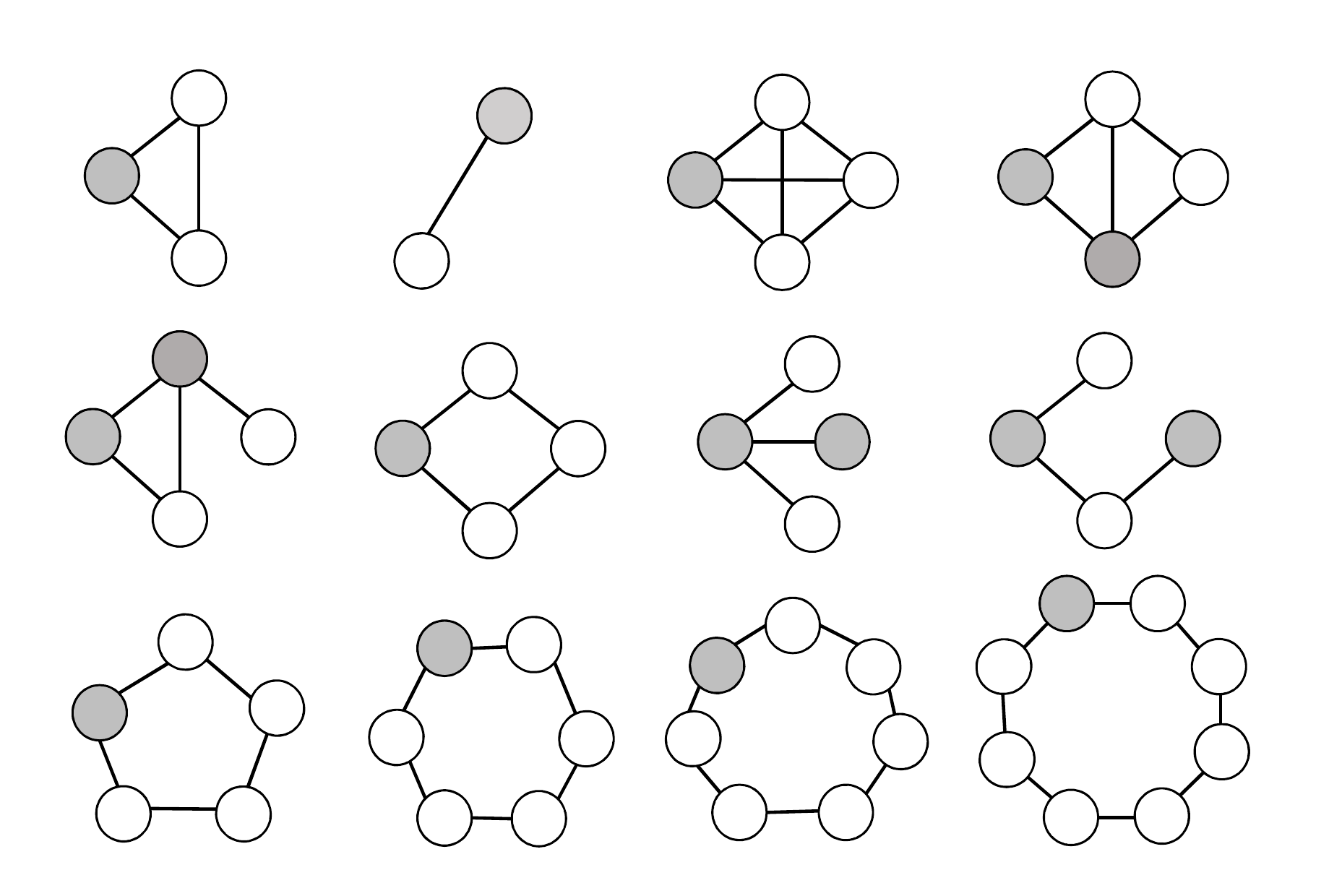}
}
\caption{\ref{fig:graphlet_vs_motif} shows the different between motif and labeled graphlet, as the former couldn't distinguish subtrees with different root. \ref{fig:graphlet_types} demonstrates all the graphlet topology in this paper.}
\end{figure*}

Graphlet is known as a representation of local structure, but existing graphlet-based statistical methods consider only topology information. Not only that, existing enumeration algorithms are limited by graphlet scale because of expensive computation costs. In attempting to solve the problems, we proposed a novel subgraph form called {\it Labeled Graphlet}. Labeled Graphlets are only marked by their starting node to simplify the complexity of the subgraph class set. As \ref{fig:graphlet_vs_motif} shows, Labeled Graphlet would tell the difference between two pairs of substructures in wireframes. Apparently, Labeled Graphlets share the same label-set $L$ with network nodes. Compared to existing graphlet statistics, our method provides additional details with little extra cost.

Generally speaking, subgraphs can be classified into three groups according to their structural characteristics: trees, paths, and circuits. Considering the variation of size, finding all subgraphs is an NP-hard problem. To simplify this procedure, we adopt twelve types of topologies as \ref{fig:graphlet_types} shows. Considering the derivation or mutually exclusive relationship between graphlets, our counting process follows the “grow” principle in a breadth-first manner. For small-scale subgraphs of size $k=3$ or $k=4$, we employ the Parallel Parameterized Graphlet Decomposition (PGD) algorithm. PGD breaks down the global searching problem into several local tasks and then merges the results for all edges. Merging and searching over low-dimensional spaces of edge neighborhoods is clearly more efficient than searching over the global high-dimensional space \cite{ahmed2015icdm}.

For the subgraphs of size $k\geq 5$, we design a dynamic programming algorithm to improve computing efficiency. The pseudocode of this process is shown in Algorithm \ref{alg:counting}. Starting from saving the first order neighbor sets of each node $a\in V$, we select tree-shaped subgraphs which meet the condition of degree. We have noticed that high-order neighbor sets can be synthesized by existing neighbor sets. For example, a 2-hop path consists of two 1-hop paths, and an 8-hop path consists of two 4-hop paths. Thus, we replace large-scale traversal with exponential steps superposition, which remarkably enhances searching efficiency with little extra space. 
The computational complexity of our algorithm is determined by line 3 of Algorithm \ref{alg:counting}, which is no more than $|V|^3$.

% \begin{breakablealgorithm}
\begin{algorithm}[h]
\label{alg:counting}
\caption{Path and circuit counting}
\KwIn{graph $G(V,E)$, maximum depth $D$, degree threshold $t$}
\KwOut{subgraph sets $Tree$, $Circuit$ and $Path$}
\textbf{Initialize}: $Circuit=Path=\emptyset$, $d$-hop adjacency matrix set $\{A^d\in \mathbb{R}^{\vert V\vert\times \vert V\vert} \vert d=2^1,2^2...2^D\}$, path set: $\{P_b(a)=\emptyset \vert\forall a,b\in V\}$\\
\For{$d\in [0,D]$}{
\For{$a,b,c \in V$}{
$N(a)=A_a^1$\\
\If{$|N(a)|\geq t$}{$Tree={N(a)+a}\cup Tree$}
\For{$i\in [2^{d}+1,2^{d+1}]$}{
\If{$A_{ab}^{(2^{d})}==1 \land A_{bc}^{(i-2^{d})}==1$}{
\If{$c \notin P_{b}(a)$}{
\If{$P_{c}(a)==\emptyset$}{$A_{ac}^{(i)}=A_{ca}^{(i)}=1$\\
$P_{c}(a)=P_{a}(c)=P_{c}(b)\cup P_{b}(a)$\\
$Path=P_{v}(a)\cup Path$\\
}
\Else{$Circuit=\{P_b(a)\cup P_c(b)\cup P_a(c)\} \cup Circuit$}
}
}   
}
}
}
\textbf{Return} $Tree, Circuit, Path$\\
\end{algorithm}
% \end{breakablealgorithm}

% \begin{breakablealgorithm}[H]
% \label{alg:counting}
% \caption{Path and circuit counting}
% \begin{algorithmic}[1]
% \STATE {//Initialization part}
%             ............
% \STATE {//Iterative part}
% \STATE  {$count\Leftarrow count+1$} 
% \UNTIL{The given termination criterion is met.}

% \end{algorithmic}
% \end{breakablealgorithm}

\subsection{Subgraph Cluster Expansion}
As mentioned above, the classical cluster expansion can be used to describe the motif structure. In order to solve this problem, we present the subgraph expansion algorithm, which omits the integral of a single node integral. In this section, we make an analogy between network subgraphs and interacting particles in the thermodynamic gas model. Thus, we can express configuration integral from the perspective of network topology, especially the network subgraph. According to this thought, Zhang et al.\cite{zhang2020graph} map the network motifs to the classical cluster expansion, then calculate the motif configuration integral and the single node integral, respectively. For graph $G$ with node set $L_v$, they re-write partition function $Z$ for the network as a sum over the individual motif contributions $z_v$,
\begin{equation}
\label{f1}
\begin{split}
    Z&=\sum \limits_v\prod \limits_{n_v}^{|L_v|}\frac{1}{n_v!}\{rq_v\}^{n_v}=\sum \limits_{n_v}z_v\\
    &=\sum \limits_{n_v} \frac{1}{n_v!}(rq_v)^{n_v}\frac{1}{N-l_vn_v}(rq_0)^{N-l_vn_v}\\
\end{split}
\end{equation}
where $r$, $n_v$, $l_v$ and $q_v$ is the radial variable, occurred frequency, number of nodes and the configuration integral (the product overall edges connecting nodes) of the $v^{th}$ motif. According to the definition presented above, motifs describe global information of networks by constructing graph representations, while graphlets attach more attention to local graph attributes. One main difference between these two statistic elements is whether the counting algorithms reuse nodes. The former one enumerates independent motifs and abandons individual nodes; The latter one count preset structures from each node while overlapping searching is inevitable. Thus, there are probably no singleton nodes left under the graphlets enumeration process. In other words, the item $N-l_vn_v$ in formula \ref{f1} can be 0 or even negative, and it becomes meaningless in the circumstances. Specifically, the configuration integral of $v^{th}$ graphlet (or the broader concept "subgraph") $q_v$ can be calculated with only connected nodes:

\begin{equation}
    q_v=\frac{1}{l_v!r}\zeta_v=\frac{1}{l_v!r}\epsilon^{d_v}
\end{equation}
$\zeta_v$ is the configuration integral obtained through the product over all edges connecting nodes, and $d_v$ is the edges number of the $v^{th}$ motif. According to the Mayer function, the configuration integral $\epsilon$ for one edge is given by
\begin{equation}
\begin{split}
    \epsilon&=\int^\infty_0(e^{-\beta v(r)}-1)dr\\
    &=\exp\left[\beta\sum\limits^{r_{max}}_{r=r_{min}}e^{-4\epsilon\left[(\frac{\sigma}{r})^{12}-(\frac{\sigma}{r})^{6}\right]}\right]+R
\end{split}
\end{equation}

where $\beta=\frac{1}{T}$ is the inverse temperature, $\sigma$ is scale parameter and $R=-\frac{r_{max}-r_{min}}{\Delta r}$. It is important to note that we confine the interval of integration to $[r_{min},r_{max}]$. The graph entropy of $v^{th}$ pattern consists of two parts: configuration integral $z_v$ and average energy $\langle U_v\rangle$.

\begin{equation}
    \begin{split}
        S_v&=\ln z_v+\beta\langle U_v\rangle\\
        &=n_v\left\{d_v\left[ \log\epsilon-\beta\frac{\epsilon-R}{\epsilon}\right]-l_v\log l_v -\log n_v \right\}\\
        %&=n_v\{ d_v\log\epsilon-l_v\log l_v-\log n_v\}+\beta\frac{n_vd_vpe^\beta}{pe^\beta+R}\\
    \end{split}
\end{equation}
where $\epsilon=pe^\beta+R$.
%the expansion formula of $v^{th}$ subgraph is

\subsection{Labeled Subgraph Entropy Kernel}
In this section, we introduce the concept of Labeled Subgraph Entropy Kernel. We replacing the occurrence frequency of graphlet with corresponding graph entropy. We provide a detailed framework description of our method and prove necessary kernel properties.
Before we define our novel kernel, we here summarize key concepts and notations. Graph $G=(V,E)$, where $V$ is a set of ordered vertices and $E\subseteq (V\times V)$ is a set of edges. If there is a mapping $V \Rightarrow L$ that assigns labels from a set $L$ to vertices, we call $G$ a labeled graph. $G$ is called undirected if $(vi, vj)\in E$ iff $(vj, vi)\in E$ otherwise, it is referred to as directed. Although many of our techniques are applicable to both directed and undirected graphs, for ease of exposition, we will exclusively deal with undirected graphs in this paper.

\textbf{Definition 1}: As for graph $G=(V,E,L)$, the corresponding labeled subgraph entropy representation around $v$ types of graphlet is
\begin{equation}
    S(G)=(S_1^1,S_2^1,...S_v^L)^T
\end{equation}

where $S_v^L$ denotes the thermodynamics entropy value measured by $v^{th}$ graphlet starting from node with label $L$. As algorithm \ref{alg:entropy} shows, $S_v^L=0$ iff the number of corresponding labeled graphlet is 0.

\textbf{Definition 2}: Let $G_1=(V_1,E_1,L)$ and $G_2=(V_2,E_2,L)$ be a pair of sample graphs. The subgraph entropy kernel $k_{SE}^{(v)}$ adopting $v$ types of graphlet is

\begin{equation}
\label{f7}
    k_{SE}^{(v)}(G_1,G_2)=k(S(G_1),S(G_2))
\end{equation}

where $k$ is the kernel function such as the Linear Product, Sigmoid Function, Polynomial Kernel Function or Radial Basis Function.

\textbf{Theorem 1}: The Labeled Subgraph Entropy Kernel $k_{SE}^{(v)}$ is symmetrical and positive definite.

\textbf{Proof}: We firstly embed two graphs into Euclidean space by formula \ref{f7}, and then calculate the kernel value using the above functions. Obviously, this process computes the inner products of two vectors in Hilbert space. Thus our Labeled Subgraph Entropy Kernel is symmetrical and positive definite.

\section{Experiment}
In this section, we demonstrate the characterization capabilities of labeled subgraph entropy kernel with several real-world datasets and perform four experiments to illustrate its multifaceted functions. On the one hand, we qualitatively analyze whether the subgraph entropy is effective in abnormal event detection of time-varying financial networks. On the other hand, we need a quantitative analysis to evaluate the effect of the proposed kernel approach. Firstly, we use subgraph entropy to evaluate the vital events within a financial time series. Secondly, we conduct Principal Component Analysis (PCA) on subgraph entropy-based graph representations and intuitively demonstrate the separation effect of our kernel. Thirdly, we perform graph classification task with subgraph entropy kernel and compare the results with several state-of-the-art methods. Finally, we design an ablation experiment to investigate the function of different physical parameters and study the effect of different topologies. The contents are as follows.

\subsection{Dataset}
\begin{table}[h]
\label{tab:datasets}
\centering
\caption{Statistical details of the datasets.}
\begin{tabular}{|l|l|l|l|l|}  
\hline  
Datasets&MUTAG~&PTC~&PROTEINS~&D\&D~\\\hline
 
Max \# nodes& 28& 109& 620&5748\\\hline
 
Mean\#nodes& 17.9& 14.69& 39.06&284.3\\\hline
 
Mean \# edges& 19.8& 14.7& 72.83&715.65\\ \hline
 
% Mean \# degree& 1.1& 1.0& 1.9&\\ \hline
 
\# graphs& 188& 344& 1113&1178\\ \hline

\# node labels& 7& 19&3&82\\ \hline
 
\# classes& 2& 2&2&2\\ \hline
\end{tabular} 
\centering
\end{table}

\subsubsection{Financial Networks}
We first validate the effectiveness of our approach on time-varying financial networks extracted from the New York Stock Exchange (NYSE) dataset\cite{silva2015modular} \cite{ye2015thermodynamic}, which contains 347 stocks with their daily closing prices over 6004 trading days from January 1986 to February 2011. In order to characterize the financial time series as a dynamic network, we use a fixed-size time window and slide it from the 29th trading day to the 6004th trading day. Each window encapsulates a set of 347 co-evolving daily stock price time series over 28 days. We compare stocks as network nodes and compute the correlation coefficients between each pair of stocks as their edge weight. Specifically, if the maximum absolute value of the weight is among the highest 5\%, we create connections between these pairs of nodes. Based on this method, we generate a family of time-varying financial networks with a fixed number of 347 vertices and varying edge structures over 5976 trading days. 
\subsubsection{Biochemistry Datasets}
The performance of the subgraph kernel has also been demonstrated on four real-world biochemistry datasets. 
Specifically, MUTAG \cite{debnath1991structure} is a collection of heteroaromatic nitro and mutagenic aromatic compounds, and the goal is to predict their mutagenicity on Salmonella typhimurium. It includes 188 samples of chemical compounds, which are labeled according to whether there is a mutagenic effect on a special bacteria—the nodes and links representing atoms and the chemical bonds, respectively.
PROTEINS is a dataset of proteins that are classified as enzymes or non-enzymes. Nodes represent the amino acids, and two nodes are connected by an edge if they are neighbors in the amino-acid sequence or in 3D space. It has three discrete labels, which represent helix, sheet, or turn.
D\&D is a dataset of 1178 protein structures obtained from \cite{dobson2003distinguishing}, classified into enzymes and non-enzymes. Each protein is represented as a graph whose nodes correspond to amino acids, and two nodes are linked by an edge if they are less than 6 Angstroms apart.
The Predictive Toxicology Challenge (PTC) dataset consists of 344 organic molecules 19 node features. Graphs are marked according to their carcinogenicity on male and female mice and rats. 
More detailed contents about the datasets are displayed in Table \ref{tab:datasets}

\subsection{Evaluation of The Entropy Time Series}
\begin{figure*}
    \label{fig:subgraph_vs_von}
    \includegraphics[width=1\textwidth]{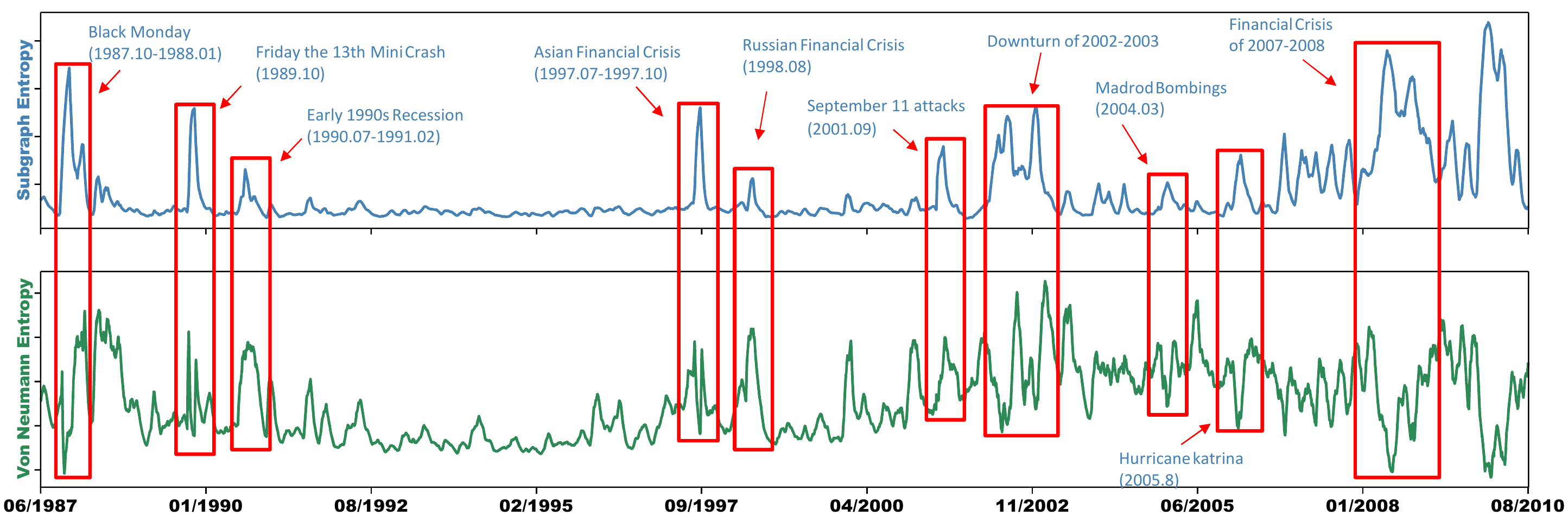}
    \caption{This figure shows the subgraph entropy of three types graphlet. The fluctuations of the first two curves are pronounced between political or economic events, while the third curve is smooth during the same period. When financial crisis happened frequently, the subgraph entropy of graphlet type 2 could be well separated, while the first two values are confusing.
}
\end{figure*}
In this experimental stage, our target is to evaluate the expressiveness of subgraph entropy on the time evolution network sequence. To analyze the time-varying financial market crisis or risk, change point detection has played an important role in identifying abrupt changes in the time series properties. For the mentioned 5976 trading days, we calculate the subgraph entropy (based on graphlet of type-2) and the Von Neumann entropy for each sample graph. According to Passerini and Severini \cite{2012The}, for a graph $G$ with adjacency matrix $A$, the von Neumann entropy of the network can be expressed as
$$S_v(G)=-Tr[\frac{\tilde{L}}{\lvert V\rvert}\ln\frac{\tilde{L}}{\lvert V\rvert} ]$$
as $\tilde{L}=D^{-1/2}(D-A)D^{-1/2}$ is the normalised Laplacian matrix, where $V$ represents the number of vertices, and $D$ is the degree matrix of $G$.

At the graph-theoretic level, this quantity may be interpreted as a measure of regularity; it tends to be larger in relation to the number of connected components, long paths, and nontrivial symmetries. Next, we use a 2-D scatter plot to show the trend of these two entropy values for the stock network, where the x-axis corresponds to the date time, and the y-axis corresponds to the subgraph entropy values and the Von Neumann entropy values. Figure \ref{fig:subgraph_vs_von} shows that both the indexes are sensitive to the financial crisis since both curves are disturbed significantly in the circles. The financial crisis includes but is not limited to Black Monday (1987.10-1988.1), Asian Financial Crisis (1997.7-1997.10), Dot-com Bubble (2000.3), Madrod Bombings (2004.3). Subgraph entropy values usually lead to a rapid increase even several days before the significant financial event. Compared to von Neumann entropy, the curve of our subgraph entropy appears smoother during normal trading days. It meant that subgraph entropy resulted in a higher recognition degree between outliers and normal trading days.

Based on the financial risk theory stated by Haubrich \cite{haubrich2013quantifying}, the financial crisis is usually caused by a set of most correlated stocks with fewer uncertainties. In other words, the financial networks are constructed by computing the correlation between pairwise stock time series, so the graphs contain a wealth of information about partial correlation (i.e., subgraph). When a financial crisis occurs, the network structure experiences dramatic changes. Therefore, our subgraph entropy values are able to promptly capture the structural changes and even provide early warning before the crisis occurs.

\subsection{Subgraph Kernel Embeddings from KPCA}
Although Fig \ref{fig:subgraph_vs_von} indicates that our subgraph entropy is effective for identifying the extreme events in the evolution of the time-varying financial networks, the indicator we adopt can only represent the network characteristics in a one-dimensional pattern space, which ignores most of within high dimensional information in Hilbert space. To better explore the performance of the proposed kernel, we perform kernel Principle Component Analysis (KPCA) \cite{witten2002data} on the kernel matrix of the financial networks and biochemistry graphs. After embedding each graph into a high-dimension vector with algorithm 1, we scatter the visualized results through the first three principal components. 

\begin{figure*}
\label{fig3}
\centering
\subfigure[Black monday]{
\label{fig:black_monday}
\includegraphics[width=0.4\textwidth]{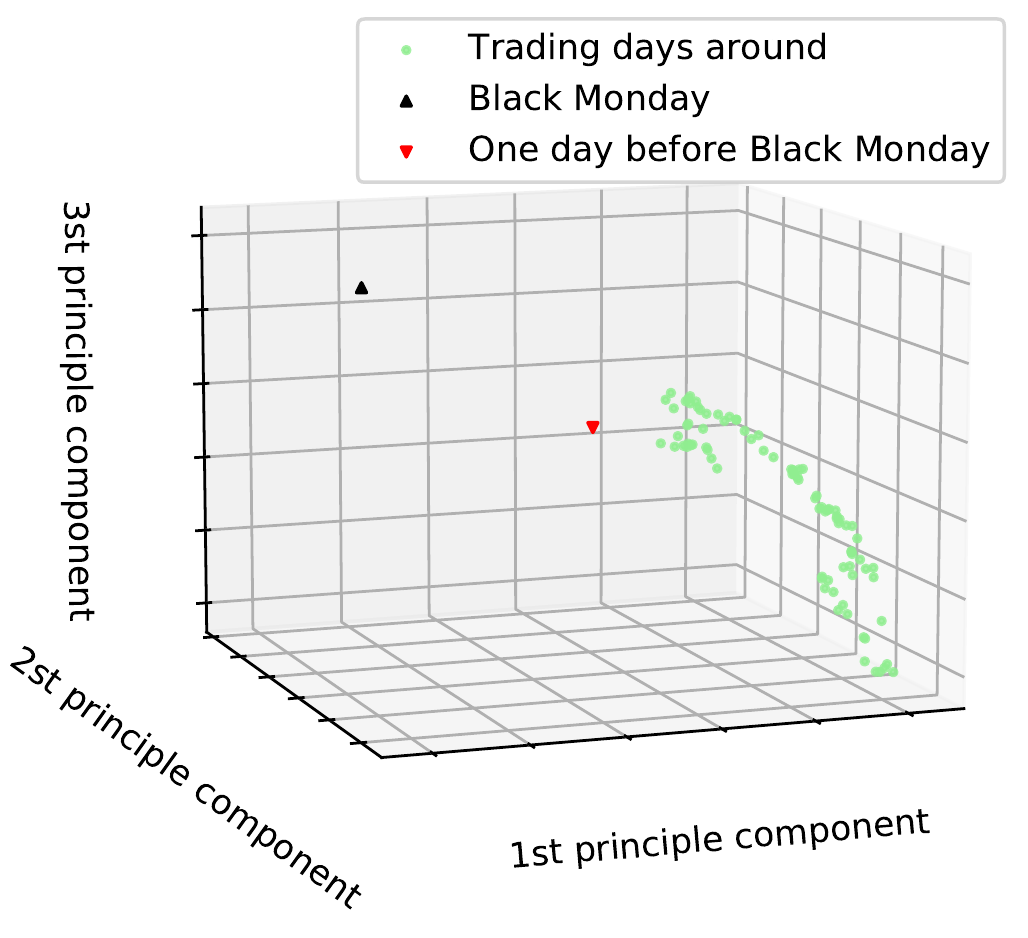}
}
\quad
\subfigure[MUTAG]{
\label{fig:mutag}
\includegraphics[width=0.4\textwidth]{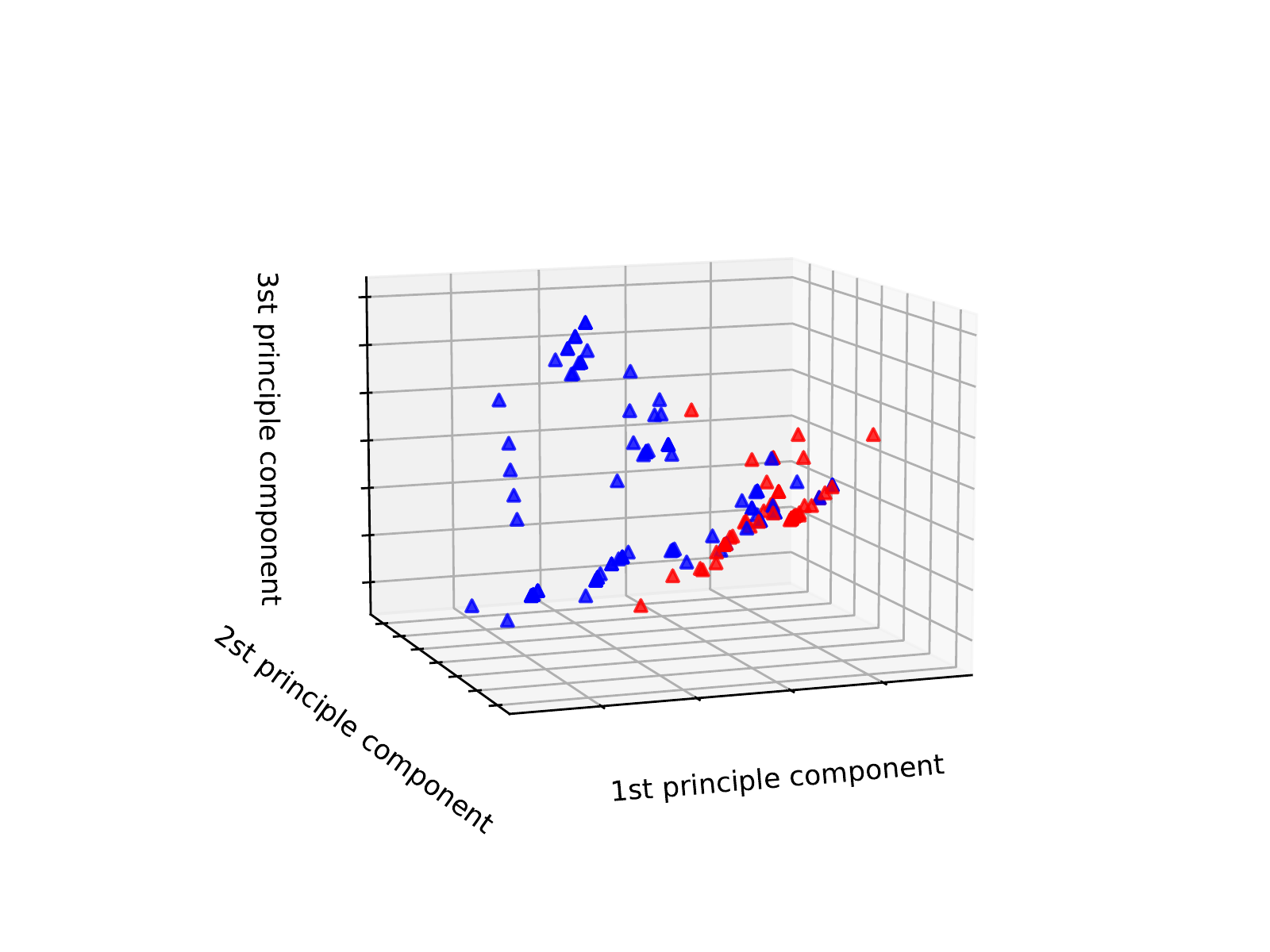}
}
\caption{\ref{fig:black_monday} demonstrates the dispersion of 90 financial networks around Black Monday. \ref{fig:mutag} shows the dispersion of all graphs within MUTAG.}
\end{figure*}

First, we scatter the trading days around a financial crisis to find more detail. The green spots respectively represent ninety trading days before and after Black Monday (October 1987). Specifically, we use a black triangle and a red inverted triangle to mark Black Monday and the day before, respectively. As Figure \ref{fig:black_monday} shows, the Black Monday and even the day before are marked as outliers, which can be detected by the first three principal components of subgraph entropy. As for the biochemistry dataset, we carry out a similar experiment. The positive and negative samples of MUTAG are marked as blue and red triangles, respectively. From Figure \ref{fig:mutag} we can see that most of the graphs are separated according to their labels, and only a few samples are mixed in the three-dimensional space. Both the results show that subgraph entropy is an effective statistical method for anomaly detection of financial and biochemistry networks.   

\subsection{C-SVM on Graph Classification}
\subsubsection{Baseline Methods}
We compare our proposed kernel LSEK with four alternative state-of-the-art methods in graph classification tasks. The methods used for comparisons include (1) Weisfeiler-Lehman subtree kernel (WL) \cite{shervashidze2011weisfeiler} with setting the depth of subtree to 2 as it could increase the feature space exponentially, (2) the attributed graph kernel from the Jensen-Tsallis $q-$differences connected with $q=2$ (JTQK) \cite{bai2014attributed}, (3) the shortest-path kernel (SPGK) \cite{borgwardt2005shortest}, (4) the graphlet count kernel (GK) \cite{shervashidze2009efficient} with setting the size of graphlets to 7 since it could exhibit the sparsity problem, (5) the shortest path kernel based on core variants (CORE SP) \cite{nikolentzos2018degeneracy}, (6) the random walk graph kernel (RWGK) \cite{kashima2003marginalized}.

\subsubsection{Classification Accuracies}
In this part, we repeat the graph classification experiment several times for each kernel and then record the average classification accuracies and standard errors. For the alternative methods, we follow the parameter setting from their original papers. For our method, we perform validation on each dataset and select the corresponding optimal hyper-parameters. Specifically, we compute the classification accuracies using the sklearn implementation of C-Support Vector Machines (C-SVM). As shown in Table \ref{tab:graph_classification}, we demonstrate the average and standard error using 10-fold cross-validation. For the sake of fairness, all the mentioned methods are run on the same computing device. 
(a)	On the MUTAG dataset, the accuracy of our LSEK evidently overcomes those of the rest of the kernels, and the CORE SP method is competitive with our method.
(b)	On the PROTEINS dataset, SEK outperforms that of all the rest of the kernels except
(c)	On the D\&D dataset, the accuracy of the LSEK method is the best. However, our method is time-consuming due to the number of vertice and node label in this dataset being rather large (some graphs contains thousands of vertices and 88 kinds of node label).
(d)	On the PTC dataset, the accuracy of LSEK surpasses the alternative kernels and is three percentage points higher than the best of the other kernels selected. Remarkably, our method clearly outperforms GK because the latter contains less local semantic information.

The experimental results show that the subgraph kernel obtains higher performance on precision at the expense of little computation time. The results suggest that our method performs better on graph classification tasks than many similar state-of-the-art kernel methods.

\begin{table}[h]
\label{tab:graph_classification}
\centering
\caption{Classification accuracy (in$\pm$standard error) }
\resizebox{\linewidth}{!}{
\begin{tabular}{|l|l|l|l|l|}
\hline  
Dataset & MUTAG & PROTEINS & D\&D & PTC \\ \hline  
WL& $82.88\pm 0.57$&$73.52\pm 0.43$&$73.39\pm 0.36$&$58.26\pm 0.47$\\ \hline
JTQK& $85.50 \pm 0.55$&$72.86 \pm 0.41$&$74.49 \pm 0.32$&$58.50 \pm 0.39$ \\ \hline
SPGK&$83.38\pm 0.81$&$75.10\pm 0.50$&$78.45\pm 0.26$&$55.52\pm 0.46$\\ \hline
GK&$81.66\pm 2.11$&$71.67\pm 0.55$&$78.45\pm 0.26$&$52.26\pm 1.41$\\ \hline
CORE SP&$88.29\pm 1.55$&$-$&$77.30\pm 0.80$&$59.06\pm 0.93$ \\ \hline
RW&$80.77\pm 0.72$&$74.20\pm 0.40$&$71.70\pm 0.47$&$55.91\pm 0.37$\\ \hline
\pmb{LSEK}&\pmb{$90.00\pm{1.66}$}&\pmb{$76.34\pm{0.68}$} & \pmb{$78.75\pm{ 0.57}$}&\pmb{$63.26\pm{1.44}$}
\\
\hline
\end{tabular}
}
\centering
\end{table}

\subsection{Parameter Setting}
\begin{figure*}
    \label{fig:ablation_financial}
    \includegraphics[width=1\textwidth]{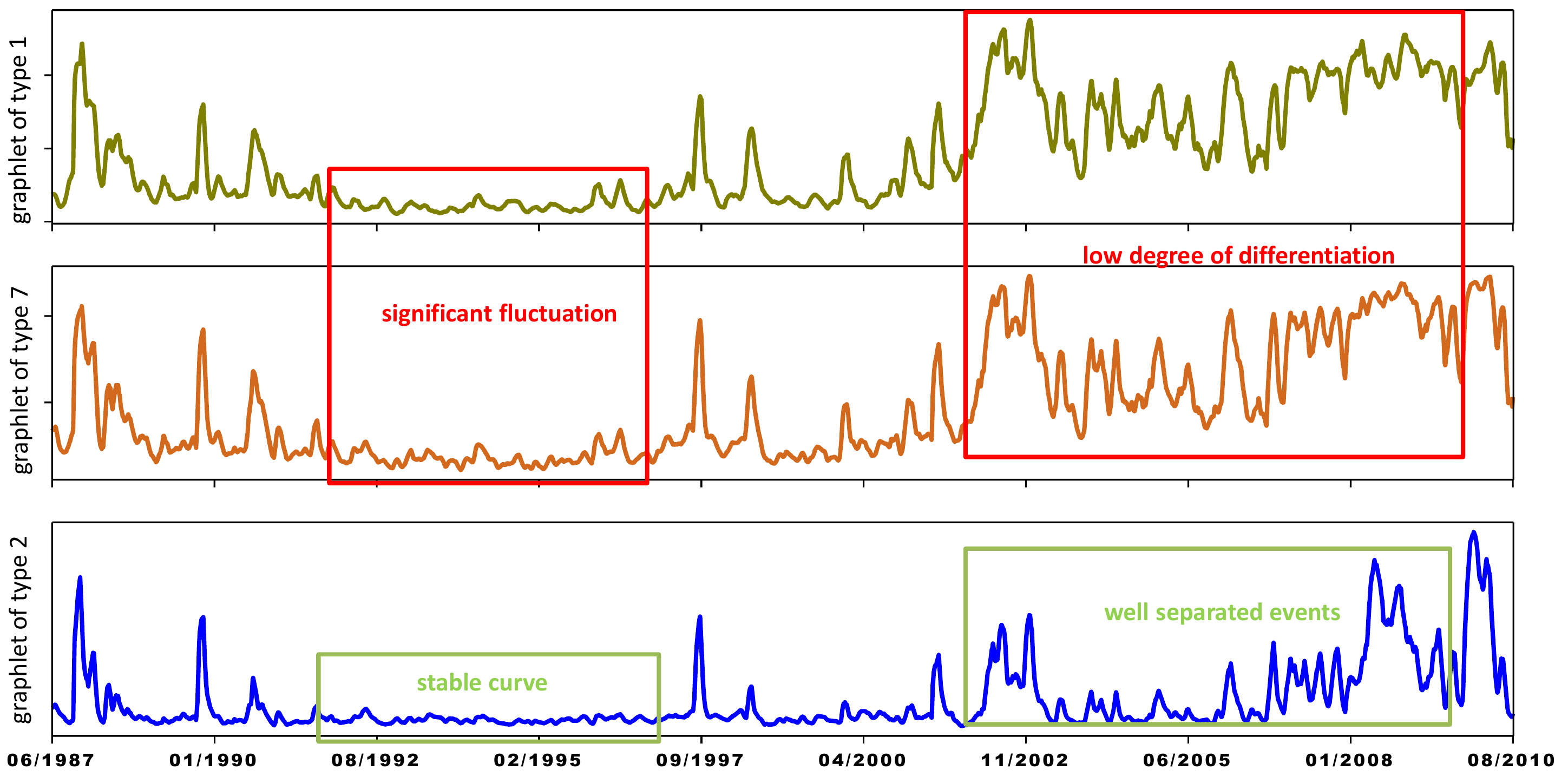}
    \caption{The characterization effect of Labeled Graphlets with different topologies.}
\end{figure*}

To explore the influence of different parameters in the subgraph kernel, we devise two ablation experiments in this subsection. Firstly, we separately record the characterization effect of different types of subgraph in mentioned time-varying financial network sequence. The view shown in Figure \ref{fig:ablation_financial} allows us to see that the curve of graphlet-type-2 fluctuates more wildly around the financial crisis, and it shows smoothness in non-crises regions. The main cause of the phenomena is that this type of graphlet appeared with relentless frequency in each trading day, while the frequency of other graphlets is much less deterministic in different situations. The result also suggests that closer relationships between the stocks are more sensitive to global political or economic events.
 
\begin{figure}
\begin{center}
\label{fig:ablation_mutag}
\includegraphics[width=0.4\textwidth]{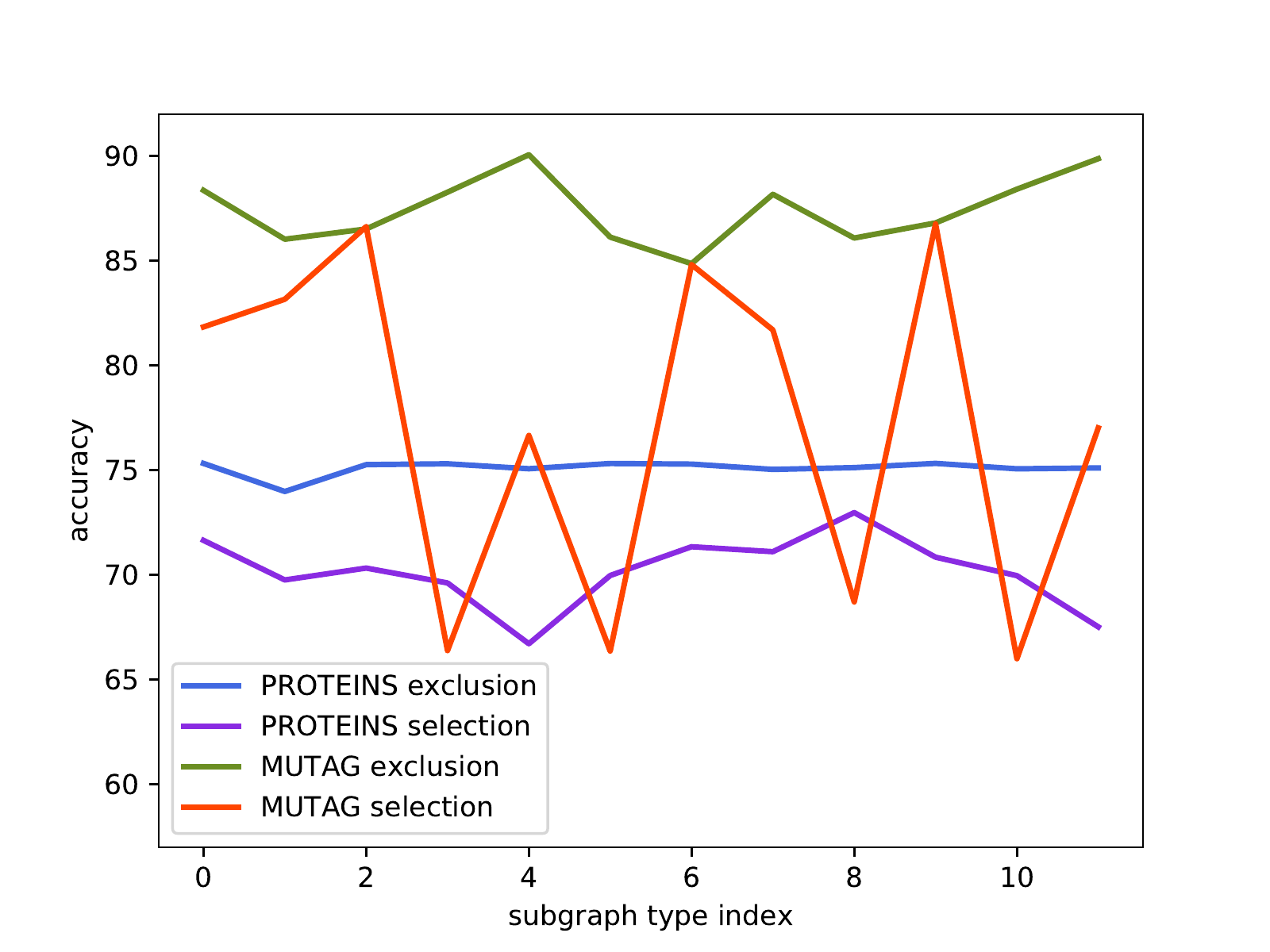}
\caption{We compare the effect of each subgraph in two datasets, and record their accuracy as these four curves above. Each point represents the corresponding effect of selecting or excluding one specific type of graphlet.}
\end{center}
\end{figure}

Then, we quantitatively analyze each mentioned subgraph with two biochemistry datasets. To better illustrate the efficiency of each topology, we design the experiment from two perspectives: exclusion and selection. The former one focuses on one specific graphlet, while the latter counts the number of all the other graphlets. As Figure \ref{fig:ablation_mutag} shows, the "exclusion" method could understandably perform to higher accuracy. Notably, the single type of graphlet also shows great characterization capability. The reason behind this phenomenon is that each labeled subgraph represents a specific chemical atomic group, which can effectively identify the graph type.

\section{Conclusion}
In this paper, we define a network index named subgraph entropy and define the corresponding graph kernel. We commerce by re-deriving the cluster expansion formula and designing a subgraph mining algorithm for labeled graphlet counting with low complexity. Driven by the combination of node feature, local structure, and global distribution, our method realizes the accurate representation of several real-world graph datasets. Our method enjoys competitive advantages with other state-of-the-art graph kernels on biochemistry datasets and demonstrates an excellent effect on financial outlier detection. Hereafter, we intend to reveal the different roles that each subgraph plays and explore the implicit relations between these subgraphs.
\section{ACKNOWLEDGMENT}
This work is supported by the Research Funds of State Grid Shaanxi Electric Power Company and State Grid Shaanxi Information and Telecommunication Company (contract no.SGSNXT00GCJS1900134), the National Natural Science Foundation of China (Grant no.61976235 and 61602535) and the Program for Innovation Research in Central University of Finance and Economics.

This work is supported by the National Natural Science Foundation of China under Grant 62176227 and U2066213.
\bibliographystyle{ieeetr}
\bibliography{ref.bib}

\end{document}